\documentclass[conference]{IEEEtran}

\IEEEoverridecommandlockouts
\usepackage{cite}
\usepackage{amsmath,amssymb,amsfonts}
\usepackage{algorithmic}
\usepackage{graphicx}
\usepackage{textcomp}
\usepackage[dvipsnames,table,svgnames,x11names]{xcolor}
\usepackage{booktabs}
\usepackage{multirow}
\usepackage{multicol}
\usepackage{tikz} % For TikZ drawing
\usepackage{lipsum} 
\usepackage{bm}
\usepackage{adjustbox}

\def\BibTeX{{\rm B\kern-.05em{\sc i\kern-.025em b}\kern-.08em
    T\kern-.1667em\lower.7ex\hbox{E}\kern-.125emX}}
\begin{document}

%\title{\modelname: Enhancing Large Vision-Language Models through Self-Reflective Sampling for Video Understanding}

\newcommand{\modelname}{Self-ReS}

\title{\modelname: Self-Reflection in Large Vision-Language Models for Long Video Understanding

\thanks{
This work is supported by NOVA LINCS ref. UIDB/04516/2020 (https://doi.org/10.54499/UIDB/04516/2020) and ref. UIDP/04516/2020 (https://doi.org/10.54499/UIDP/04516/2020) with the financial support of FCT.IP; and Fundação para a Ciência e Tecnologia ref. 2023.03647.BDANA.}
}
\vspace{-0.2em}
\author{
  João Pereira$^{1,2}$,
  Vasco Lopes$^{2,3}$,
  David Semedo$^{1}$,
  João Neves$^{3}$\\
  $^{1}$NOVA LINCS, NOVA@FCT, Caparica, Portugal.
  \\
  $^{2}$DeepNeuronic, Covilhã, Portugal. 
  \\
  $^{3}$NOVA LINCS, University of Beira Interior, Covilhã, Portugal. \\
  jaca.pereira@campus.fct.unl.pt
  \vspace{-1.2em}
}

\maketitle

\begin{abstract}

Large Vision-Language Models (LVLMs) demonstrate remarkable performance in short-video tasks such as video question answering, but struggle in long-video understanding. The linear frame sampling strategy, conventionally used by LVLMs, fails to account for the non-linear distribution of key events in video data, often introducing redundant or irrelevant information in longer contexts while risking the omission of critical events in shorter ones. To address this, we propose \modelname, a non-linear spatiotemporal self-reflective sampling method that dynamically selects key video fragments based on user prompts. Unlike prior approaches, \modelname~leverages the inherently sparse attention maps of LVLMs to define reflection tokens, enabling relevance-aware token selection without requiring additional training or external modules. Experiments demonstrate that \modelname~can be seamlessly integrated into strong base LVLMs, improving long-video task accuracy and achieving up to 46\% faster inference speed within the same GPU memory budget. The code can be found here: https://github.com/jaca-pereira/Self-ReS.
\end{abstract}


\vspace{-1em}
\section{Introduction}

Recently, Large Vision-Language Models (LVLMs) emerged as a way to leverage the strong zero- and few-shot performance of Large Language Models (LLMs) in downstream vision tasks. These models combine vision and language by projecting visual features from a vision encoder into the LLM's embedding space, followed by fine-tuning on multimodal data. In the video domain, the adoption of strong backbone models~\cite{siglip}\cite{qwen2} and the development of high-quality pre-training and instruction-tuning datasets \cite{LLaVA-Video} has led to improved performance on tasks such as video question answering \cite{VideoMME}. While LVLMs demonstrate strong capabilities in understanding short videos, their performance on longer videos remains a significant challenge, due to difficulties in modeling larger numbers of frames. In the reference benchmark Video-MME [4], several state-of-the-art models such as LLaVA-Video [3] report strong performance for short videos, but struggle to obtain competitive performance for longer inputs, especially without the textual hints present in subtitles. 
%This is particularly evident in the mid-sized (7B) variants, where accuracy can drop by approximately 20\%.

Efforts to improve performance for longer videos can be categorized into two main approaches: a) increasing context size, and b) applying token compression techniques. Often, models propose a combination of the two. 
Despite significantly increasing the number of sampled frames - e.g., up to 768 for Qwen2-VL~\cite{Qwen2-VL} and 1024 for NVILA~\cite{NVILA}- and applying aggressive token compression to minimize the cost of reasoning over larger contexts, these models still struggle with long videos, being unable to outperform models with significantly smaller context sizes - e.g., LLaVA-Video with 64 frames only

\begin{figure}[t]
    \centering
    \includegraphics[width=\linewidth]{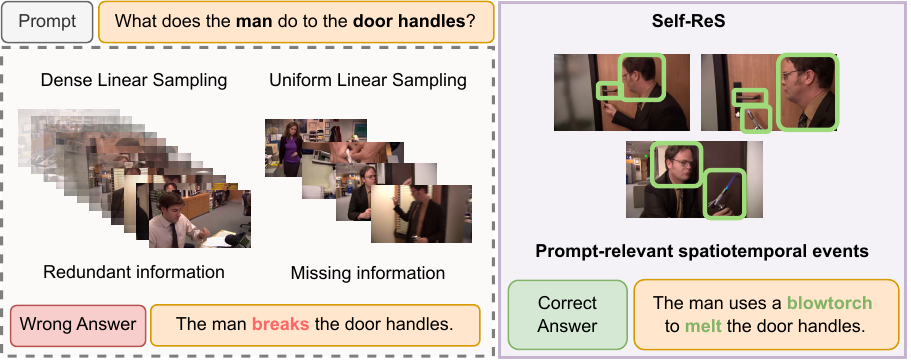}
    \caption{\textbf{Linear Frame Sampling vs Self-Reflective Sampling (\modelname{}).}
    The linear frame sampling method, used by most LVLMs, samples a dense or uniform subset of frames, increasing the likelihood of sampling redundant and background information or missing key events. In contrast, \modelname{} increases sampling granularity, optimizing context-window usage with prompt-relevant fragments of the video identified through spatiotemporal self-reflection.}
    \label{fig:teaser}
    \vspace{-1em}
\end{figure}

We argue that current frame sampling strategies overlook video non-linearity, where critical events, such as a sudden action or a key object appearance, occur in a concentrated spatiotemporal region, rather than being evenly distributed throughout the video - e.g., Figure \ref{fig:teaser}.  Moreover, linearly scaling the number of frames to accommodate longer videos will likely introduce background and redundant information, inefficiently allocating more computational resources. 

An important line of research focuses on proposing improved sampling strategies for video LVLMs. Widely adopted techniques such as token pooling \cite{VideoChatGPT}~\cite{NVILA}~or merging~\cite{ChatUniVi} allow for extended context sizes, but also fail to account for the non-linear nature of video data. More recent approaches attempt to address this non-linearity through dynamic memory banks~\cite{MovieChat}, adaptive token compression ~\cite{LongVU}\cite{oryx} or by learning a summarization token~\cite{VideoXL}. Notably, the proposed solutions rely on external, trainable modules, introducing added computational requirements that may affect their adoption in real-world scenarios.

In this paper, we introduce \textbf{\modelname}, a novel \textit{Self-Reflective Sampling} approach designed to enhance LVLM performance on long video understanding. \modelname~addresses video non-linearity by selecting key spatiotemporal fragments, guided by textual cues present in the user prompt. Our approach builds on the idea that LLMs can self-reflect and critique their own generations \cite{selfreflectionrag}~to improve their final answer. In particular, \modelname~leverages the inherent self-reflective ability of base LLMs to identify relevant video fragments according to a user-specified task.

\modelname~introduces independent reflection paths, where each path processes, in parallel, a different segment of the video. These paths converge at specific depths within the LLM, enabling unified inference across the entire video. The self-reflective nature of the LLM is expressed in \textit{reflection tokens}, which serve as judges to evaluate the relevancy of visual tokens based on the user query. By leveraging the inherent ability of LVLMs to generate sparse attention maps\cite{FastV}, which naturally focus on a subset of visual tokens, \modelname~defines reflection tokens to prioritize critical information. Once the reflection paths converge, the LLM performs inference using its default context size, concentrating on the most salient visual tokens, which are non-linearly selected according to the textual cues in the user prompt. With this strategy, \modelname~achieves up to $\mathbf{46}\textbf{\%}$ faster inference speed and higher accuracy than the base model while maintaining GPU memory budget. 

Our contributions can be summarized as follows:
\begin{itemize}
    \item We propose \textbf{\modelname}, a \textit{Self-Reflective Sampling} method that dynamically identifies and prioritizes key spatiotemporal fragments of video data. By leveraging the inherent self-reflective capabilities of LLMs, \modelname~addresses spatiotemporal non-linearity without requiring additional training or external modules.

    \item \modelname~enables LVLMs to process significantly more frames while reducing computational overhead. It achieves higher accuracy with up to \textbf{46\% faster inference speed} and identical GPU memory usage, making it practical for resource-constrained scenarios.

    \item Using \modelname, LLaVA-Video outperforms existing open and closed-source models on two reference video benchmarks.
\end{itemize}

\section{Related Work}

\subsection{Large Vision-Language Models}

LVLMs extend the zero-shot and instruction-following capabilities of LLMs to visual inputs, enabling multimodal tasks across images and videos. Foundational LVLMs such as LLaVA~\cite{LLaVA} and BLIP~\cite{BLIP2} connect vision encoders to LLMs via linear projections or learned queries, forming the basis for pioneer video-specific models such as VideoChatGPT~\cite{VideoChatGPT}. These models are pre-trained on video-captioning and instruction-tuning datasets~\cite{LLaVA-Video}, enabling strong performance in short video tasks.

Recent advancements scale LVLMs to longer videos using state-of-the-art backbones~\cite{siglip}\cite{qwen2} and increased frame context sizes~\cite{NVILA}\cite{Qwen2-VL}. However, these improvements often come at the cost of significant computational overhead. Moreover, these methods fail to address spatiotemporal non-linearity in video data, where key events are concentrated in small regions rather than distributed uniformly. \modelname~enables foundational models to process longer videos within the default context size, dynamically prioritizing salient tokens while maintaining computational efficiency.

\subsection{Video Sampling Strategies for LVLMs}

Sampling strategies are critical for extending LVLM performance to longer videos. Early approaches such as VideoChatGPT~\cite{VideoChatGPT} and ChatUniVi~\cite{ChatUniVi} employed token pooling or merging to handle the larger number of video frames. Memory-based strategies such as MovieChat~\cite{MovieChat} store long context video information in memory tokens. LLaMA-ViD~\cite{li2024llamavid} learns a single token per frame. Recent proposals include learnable video summarization tokens in VideoXL~\cite{VideoXL} and spatiotemporal adaptive token compression in LongVU~\cite{LongVU}, which uses high-resolution vision encoders to compute frame similarity and cross-modal relevance. While effective to an extent, these methods often rely on additional backbones and require extensive training \cite{LongVU}\cite{VideoXL}\cite{li2024llamavid}, increasing computational costs. Moreover, they may fail to adequately address video non-linearity \cite{MovieChat}\cite{VideoChatGPT}\cite{ChatUniVi}, where key events occur in sparse spatiotemporal fragments of the video. In contrast, \modelname~leverages the LLM’s self-reflective abilities, expressed in the form of native attention weights that dynamically identify and prioritize critical visual tokens. Our plug-and-play approach eliminates the need for external modules or additional training, offering a computationally efficient solution that is faster than the baseline and surpasses prior methods in accuracy.

\section{Methodology}

\begin{figure}[t!]
    \centering
    \includegraphics[width=\linewidth]{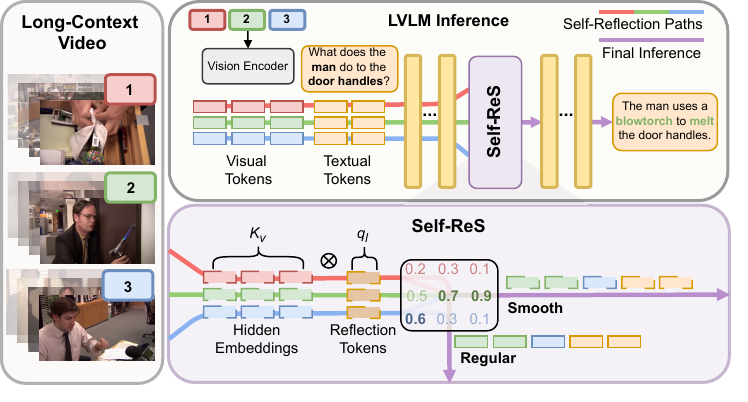}
    \caption{\textbf{Illustration of \modelname~methodology.} \modelname~splits the long-context video into reflection paths, performing batch inference to a specific depth. Reflection tokens then assess the relevancy of visual tokens for the user task. After self-reflection, the paths merge into a single inference path using either Regular or Smooth Inference modes. Regular mode samples salient tokens and restarts the forward pass from the first layer, while Smooth mode continues from the self-reflection layer with the most salient hidden embeddings.}
    \label{fig:arch}
    \vspace{-1em}
\end{figure}

In this section, we detail the methodology behind \modelname, illustrated in Figure \ref{fig:arch}. \modelname~is designed to address three key challenges faced by LVLMs in long video understanding: 1) \textbf{Non-Linearity of video data}, where critical events, such as short actions or key object appearances, are often concentrated in small spatial and temporal regions of a video, rather than being evenly distributed; 2) \textbf{Limited context sizes} which restrict the number of frames that can be processed, risking the omission of key events in shorter temporal windows; 3) \textbf{Linearly increasing context size} introduces \textbf{background noise and redundant information}, leading to inefficient resource utilization.

\modelname~addresses these challenges through a process called \textbf{spatiotemporal self-reflection}. This process begins by sampling a large number of frames from a long video and distributing them across parallel reflection paths. Each path processes a segment of the video independently up to a specified layer depth in the LLM. The paths are then merged into a single inference path, where the most salient visual tokens are non-linearly selected based on the user task. By dynamically prioritizing key information and minimizing redundancy, \modelname~effectively expands the context size without inefficiently scaling computational resources.

\subsection{Spatiotemporal Self-Reflection}

Inspired by the concept that \textbf{LLMs can critique their own predictions} through parallel and independent inferences~\cite{selfreflectionrag}, we design reflection tokens to judge the relevancy of visual tokens for a given task. Our approach leverages the observation that the \textbf{last input token of LVLMs inherently generates sparse attention maps for visual tokens}~\cite{FastV}. By adopting the last input token as the reflection token, \modelname~avoids the need for additional training.

Without loss of generality, we assume a standard LVLM architecture composed of a vision encoder $f_V$, a linear projection matrix $W\in \mathbb{R}^{d_V \times d}$, and an LLM $f$, where $d_V$ and $d$ correspond to the visual encoder and LLM token dimensionality, respectively. We also assume the default input sequence formatting, with a system prompt $X_{\text{sys}}$ being concatenated with the video features, followed by the user question $X_{q}$.

To address the context size limitations, \modelname~emulates a larger context size by sampling $T$ frames from a long video and distributing them across $R$ independent reflection paths. Each path processes $S=\frac{T}{R}$ frames, where $S$ matches the base model's default context size. The frames are first processed by the Vision Encoder $f_V$, generating visual features $X_V \in \mathbb{R}^{R \times N_V \times d_V}$, where $N_V$ represents the number of visual tokens per reflection path. These features are then projected into the LLM embedding space using the matrix $W$, producing $\hat{X}_V \in \mathbb{R}^{R \times N_V \times d}$. The inputs for each reflection path $r_i$ are defined as:
\begin{equation}
X_{r_i} = [X_{\text{sys}} \oplus \hat{X}_V^i \oplus X_{q}],
\end{equation}
with $X_{r_i} \in \mathbb{R}^{ N_r \times d}$, where $\hat{X}_V^i$ corresponds to the projected visual tokens of path $r_i$, $N_r$ is the total number of tokens per reflection path, and $\oplus$ denotes concatenation.

Each reflection path sequence $X_{r_i}$ is forwarded independently, thus having its own attention mask and position embeddings. Depending on the available memory resources, \modelname~can operate in parallel (batch mode) or sequentially (memory-efficient mode), ensuring scalability across diverse computational environments. For simplicity, we assume parallel inference mode in this explanation.

Our self-reflective approach begins with the LLM's prefill stage, performing a forward pass up to a designated self-reflection layer. At this layer, \modelname~applies \textbf{Self-Reflective Attention (SRA)}, a modified version of Scaled Dot-Product Attention designed to perform a non-linear selection of visual tokens. This mechanism determines query-key similarities without the memory-intensive value multiplication and softmax operations. Also, introducing a new attention mechanism only at the self-reflection layer allows compatibility with all existing LVLM setups, including those using Flash Attention \cite{flash-attn}, which typically does not expose attention weights.

Specifically, we designate the hidden embedding of the last token of the user query $X_{q}$, at self-reflection layer $l$, as our reflection token. The reflection token is then projected as the query vector $q_l \in \mathbb{R}^{N_V \times d_q}$. Likewise, the hidden embeddings of the visual tokens are projected into the key matrix $K_{V_l}\in \mathbb{R}^{N_V \times d_k}$, where $d_k=d_q$. SRA is computed as:

\begin{equation}
\label{eq:selfreflective}
\text{SRA}(q_l, K_{V_l})=\frac{q_l \cdot K^T_{V_l}}{\sqrt{d_k}}.
\end{equation}
We purposely omit the attention head dimension for simplicity. The final saliency scores are computed by averaging the scores for each token over all attention heads, isolating individual contributions.

% O(S*(N*N)) << O(N_large * N_large)

%é mais facil ter a complexidade em relação ao número de frames, pois é linear ao número de frames
%footnote - sendo que numero de tokens de 1 frame é -letra-, o O(letra*N_v*S) =  =(N_v*S)
%O "O" da complexidade não é este!!!!!

%mencionar também a complexidade global da solução
%a complexidade com a nossa solução é em teoria ligeiramente menor

%comparar tempo de inferência de
% num só batch
% em batch
% com menos frames
%a complexidade continua quadrática mas é menor

%O(128*128) -> O(#LayersFull * (N*N))
% 
% w/ Self-Res: O(#LayersPartial * k * ((N/k)*(N/k)) + O(#LayersFull * (N*N)/k)
% w/ Self-Res: O(#LayersPartial * (N)/k) + O(#LayersFull * (N*N)/k^2)

% w/ Self-Res: O((N*N)/k) + (N*N)/k^2) -> Majorando: O(N*N)/k)
% # LayersPartial = #LayersFull / 5
% w/ Self-Res: O((N*N)/k) + (N*N)/k^2) -> Majorando: O((#LayersFull / 5) * N*N*(1/k))
%O(64*64)
%O(2*(64*64)) O((N/2)² * (N/2)²) 

The last stage of spatiotemporal self-reflection utilizes the judgments produced by SRA to evaluate the relevancy of visual tokens across reflection paths. These judgments are derived from the scores calculated from the sparse attention maps of the reflection tokens. The expressiveness of these scores arises from the fact that reflection tokens $q_l$ act as highly informational embeddings for next-token prediction \cite{FastV}.

Once the attention weights are calculated, \modelname~selects the top-$N_V$ visual tokens:
\begin{equation}
S_c = \text{Top}_{N_v} \left( \{\text{SRA}_{r_i}(q_l, K_{V_l})\}_{r_i=1}^{r_i=R} \right),
\end{equation}
where $S_c$ contains the indices of the most salient visual tokens over all reflection paths. This selection process introduces non-linearity in both the spatial and temporal dimensions and eliminates background information, retaining only the most critical tokens. The stored indices in $S_c$ are then used to sample tokens in the next phase, where the reflection paths converge into a unified inference path.

\subsection{Converging Reflection Paths}

The final stage of \modelname~merges the reflection paths into a single inference path, enabling the model to operate within its default context size while reasoning over the non-linear salient video fragments selected during spatiotemporal self-reflection. This step allows the model to effectively capture relationships across video segments. To achieve this, \modelname~implements two complementary strategies for converging reflection paths, tailored to balance computational efficiency and full model reasoning.

\paragraph{Regular Inference} This mode initiates a new forward pass from the first layer using only the top-$N_v$ visual tokens selected during the spatiotemporal self-reflection phase. The input sequence is constructed as:
\begin{equation}
X = [X_{\text{sys}} \oplus \hat{X}_r \oplus X_{\text{q}}],
\end{equation}
where $\hat{X}_r \in \mathbb{R}^{N_v \times d}$ represents the set of projected visual tokens with the highest scores $S_c$, from the flattened sequence of visual tokens across all reflection paths. The original temporal order of the visual tokens is maintained, enabling the LLM to perform a full inference pass on the selected salient tokens.

\begin{figure}[t]
    \centering
    \includegraphics[trim={0 20pt 0 0},clip,width=\linewidth]{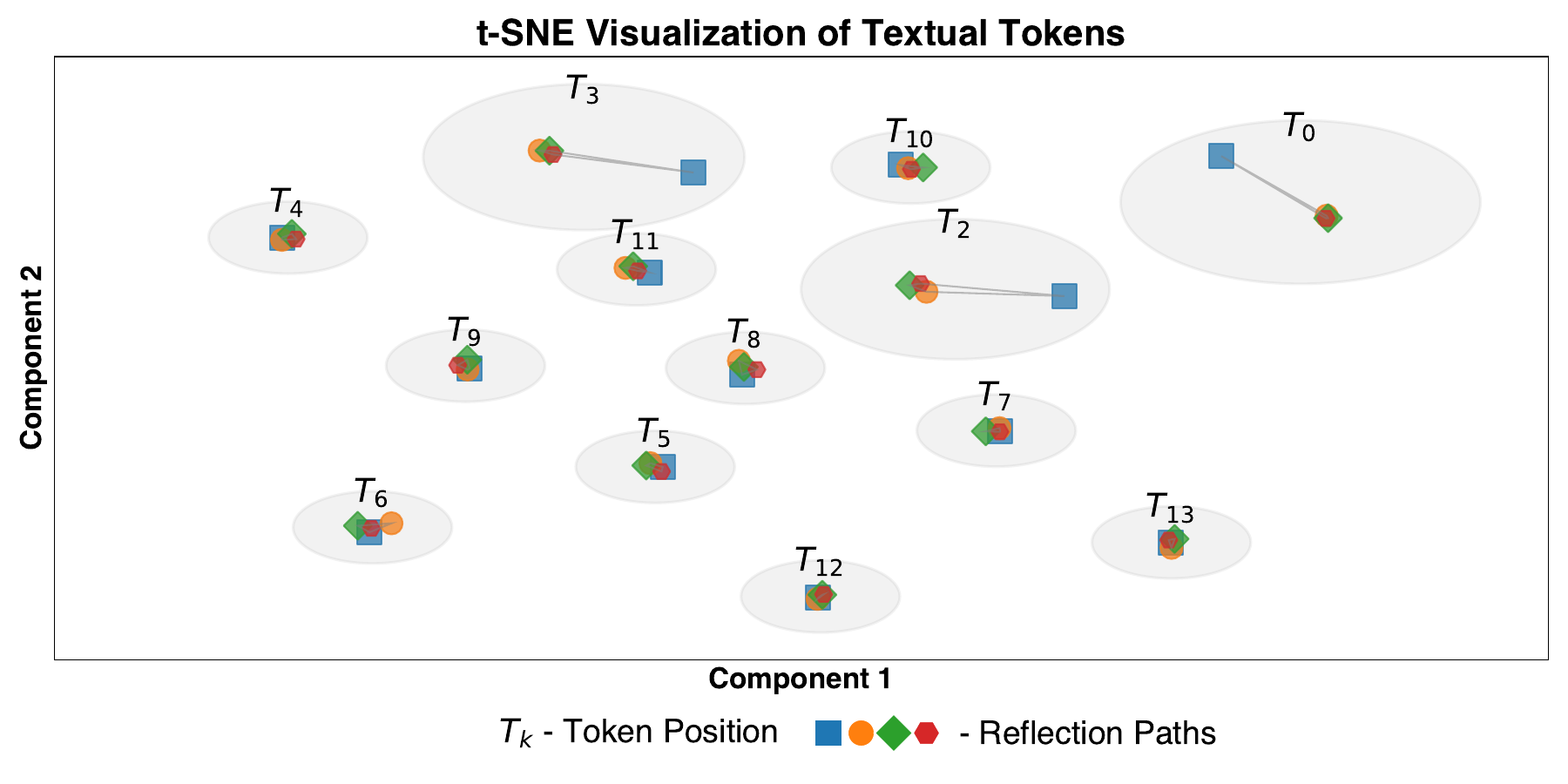}
    \caption{\textbf{Consistency of system and textual hidden representations over reflection paths.} The hidden representations of the system and textual input have been reduced to a low-dimensional space (2D), evidencing the residual difference between the hidden representations over different reflection paths.}
    \label{fig:tsne-hs}
    \vspace{-1em}
\end{figure}

\paragraph{Smooth Inference} This mode leverages the hidden embeddings after the Self-Reflection layer $l$, and continues the forward pass with the top-$N_v$ visual tokens, instead of restarting from the first layer. We define the input sequence for the final inference path as:
\begin{equation}
H = [H_{\text{sys}_0} \oplus H_r \oplus H_{\text{q}_0}],
\end{equation}
where $H_{\text{sys}_0}$ and $H_{\text{q}_0}$ are the hidden embeddings of the system and user tokens from the first reflection path, and $H_r \in \mathbb{R}^{N_v \times d}$ represents the top-$N_v$ self-reflected hidden embeddings of salient visual tokens, conditioned on the user query across all reflection paths. As shown in Figure \ref{fig:tsne-hs}, the hidden embeddings of text tokens generated with the same prompt across different reflection paths converge to equivalent representations. We wield this insight to define our smooth converging strategy and select the textual tokens of a single reflection path. Smooth Inference minimizes redundant computations, maintaining high inference efficiency. 

\paragraph{Final Inference Path} Regardless of the chosen strategy, \modelname~converges into a unified new inference path sequence. This sequence is forwarded using vanilla attention and reassigned positional embeddings (see section \ref{app:pos_embbs} of the Appendix). This path focuses exclusively on the most salient visual tokens, identified through the LLM’s self-reflective capabilities guided by the user query. By leveraging these tokens, the final model prediction effectively captures global inter-segment dependencies, enabling reasoning over a larger context of the input video. Notably, this is achieved without actually increasing the model’s default context size or memory requirements, ensuring efficient handling of non-linear video events.

\subsection{Complexity of \modelname}

Our proposed SRA effectively computes saliency scores for visual tokens, similar to regular attention mechanisms but with reduced computational overhead. In our baseline LVLM, the computational cost of attention is quadratic over the total number of input tokens - $O(R*N_r^2)$, where $R$ is the number of reflection paths and $N_r$ is the number of input tokens per reflection path. In \modelname, while attention computation remains quadratic, it is computed over a smaller number of input tokens split between reflection paths: $O(R*N_r^2) = O(R*(N/R)^2) = O(N^2/R)$. Thus, while \modelname~maintains quadratic complexity, it is reduced by a factor of $R$. SRA (Eq.~\ref{eq:selfreflective}) has linear complexity $O(N_v)$, which is then applied to each reflection path $O(R\cdot N_V)$. Therefore, it does not affect the overall complexity of the LVLM inference flow.

\section{Experiments}
%1,5 pages for experiments

In this section, we thoroughly evaluate the impact of integrating \modelname~in a state-of-the-art foundational LVLM using two reference video understanding benchmarks. We benchmark \modelname~against existing foundational LVLMs and state-of-the-art sampling methods to demonstrate its effectiveness in improving accuracy and efficiency.

\subsection{Datasets.} 
\paragraph{Video-MME~\cite{VideoMME}} Encompasses multiple choice questions over challenging short, medium and long videos. The videos cover a wide range of topics including Knowledge, Film \& Television, Sports Competition, Artistic Performance, Life
Record, and Multilingual. We focus on the benchmark without subtitles because it closely simulates real-world scenarios where subtitles are unavailable, and ensures that the evaluation better reflects the model's ability to comprehend visual spatiotemporal information.

\paragraph{MLVU~\cite{mlvu}} The Multi-task Long Video Understanding benchmark is used to evaluate performance of the proposed strategy on diverse and complex long-video understanding tasks. MLVU comprises a collection of videos ranging from 3 minutes to 2 hours, spanning diverse genres such as movies, documentaries, and surveillance footage. The benchmark features nine distinct task categories, including topic reasoning, anomaly recognition, summarization, and action understanding, designed to test holistic, single-detail, and multi-detail video comprehension.

\subsection{Implementation Details}

For all experiments, we use a greedy decoding strategy. These values ensure deterministic and reproducible results. For \modelname~specific parameters, we explore different combinations of total frames sampled ($T$), segment size ($S$), and the self-reflection layer depth, as detailed in the ablation study. We rely on an open-source library~\cite{transformers} to implement SRA at a designated self-reflection layer and batch processing for reflection paths.

\subsection{Long Video Understanding}
Table \ref{tab:results} compares the performance of \modelname~against other LVLMs with different sampling strategies. Results clearly demonstrate the superiority of \modelname~in comparison to methods using similar sized LLMs, achieving state-of-the-art results. LLaVA-Video equipped with \modelname~surpasses both the base model and models with much larger context sizes. Additionally, our spatiotemporal self-reflective sampling methodology helps LLaVA-Video outperform much larger closed and open source methods, including GPT-4o~\cite{GPT4o} and GPT-4o mini~\cite{GPT4o-mini}.

These results evidence the benefit of non-linear token selection at both the spatial and temporal dimensions of the video. \modelname~uses non-linear spatiotemporal self-reflection to compress task-relevant information into a significantly smaller context window, contrasting with current approaches that linearly extend the context across numerous frames, leading to redundancy and background noise. Furthermore, our training-free approach, leveraging the naturally sparse attention maps of the last user query token, allows for our method to surpass other trainable, non-linear sampling methods, by building on and improving a strong base model. 

\begin{table}
    \centering
    \caption{Performance comparison on Video-MME and MLVU\textsuperscript{\textdagger}. }
    \resizebox{.5\textwidth}{!}{
    \begin{tabular}{llllccc}
    \toprule
       & \textbf{Model} & \centering\textbf{Sampling} & $\bm{T}$ & \multicolumn{2}{c}{\textbf{Video-MME}} & \textbf{MLVU}\\ 
        & & \centering\textbf{Strategy} &  & Long   & All & \\ 
    \midrule
    \multirow{4}{*}{\rotatebox{90}{\shortstack{Large\\$>=$72B}}}&VideoLLaMA2\cite{VideoLLaMA2} & Q-Former & 32 & 57.6 & 62.4 & 45.6\\ 
     &LLaVA-OV\cite{LLaVA-OneVision} & Pooling& 32 & 60.0 & 66.3 & 68.0\\
     &GPT-4o mini\cite{GPT4o-mini}& - & 250 & 58.6 & 64.8 & -\\ 
     &GPT-4o\cite{GPT4o} & - & 384 & 65.3 & 71.9 & 64.6\\ 
    \midrule
    %ChatUniVI, LLaVA PruMerge
        \multirow{9}{*}{\rotatebox{90}{\shortstack{Mid-Sized\\7B}}}&MovieChat\cite{MovieChat} & Memory & 2048 & - & - & 18.0 \\
        &LLaMA-ViD\cite{li2024llamavid} & CrossAt & 1fps & - & - & 17.2 \\
        &ChatUniVI\cite{ChatUniVi} & Merging & 64 & 35.8 & 40.6 & - \\
        &VideoXL\cite{VideoXL} & VST & 2048 & 55.5 & 61.0 & 64.9 \\
        &LongVU\cite{LongVU} & SVA  & 1fps  & - & - & 65.4 \\
        &Oryx\cite{oryx} & CrossAt & 128 & - & 58.8 & 67.5 \\
        &Qwen2-VL\cite{Qwen2-VL} & MLP & 738 & -& 63.3 & -\\
        &NVILA\cite{NVILA} & Pooling & 1024  & 54.8 & 64.2 & 70.1 \\
        %& \textcolor{gray}{LLaVA-Video$^{*}$} & \textcolor{gray}{Pooling} & \textcolor{gray}{64} & \textcolor{gray}{-} & \textcolor{gray}{63.3} & \textcolor{gray}{70.8} \\

     \midrule
        &\multirow{3}{*}{LLaVA-Video$^{}$\cite{LLaVA-Video}} &\multirow{3}{*}{Pooling} & 32 & 52.2& 62.5 & 66.5\\
                                            & &                            & 64 &53.4 & 64.3 & 70.3\\
                                            & &                            & 128 & 54.1 & 64.6 & 70.3\\
    \midrule
        &\multirow{2}{*}{LLaVA-Video} & \multirow{2}{*}{\textbf{{\modelname}}}  & 128 (32) & 54.6 & 64.3 & 67.9\\ 
                                     & &                                          & 128 (64) & \textbf{{56.3}} & \textbf{{65.5}} & \textbf{{70.7}}\\

    \bottomrule
    \end{tabular}
    }
    \label{tab:results}
    \centering{\\
      \textsuperscript{\textdagger} $\bm{T}$ denotes the total number of frames sampled for input. VST is Video Summarization Token, SVA is Spatiotemporal Visual Aggregation and CrossAt is Cross-Attention. '-' means value not reported.  
    }
\vspace{-10pt}
\end{table}

\subsection{Ablation Study}

\begin{figure}[t!]

\begin{tikzpicture}
    % Include the image
    \node at (0, 0) {\includegraphics[width=0.32\textwidth, trim={0 12pt 0 0},clip, height=3.25cm]{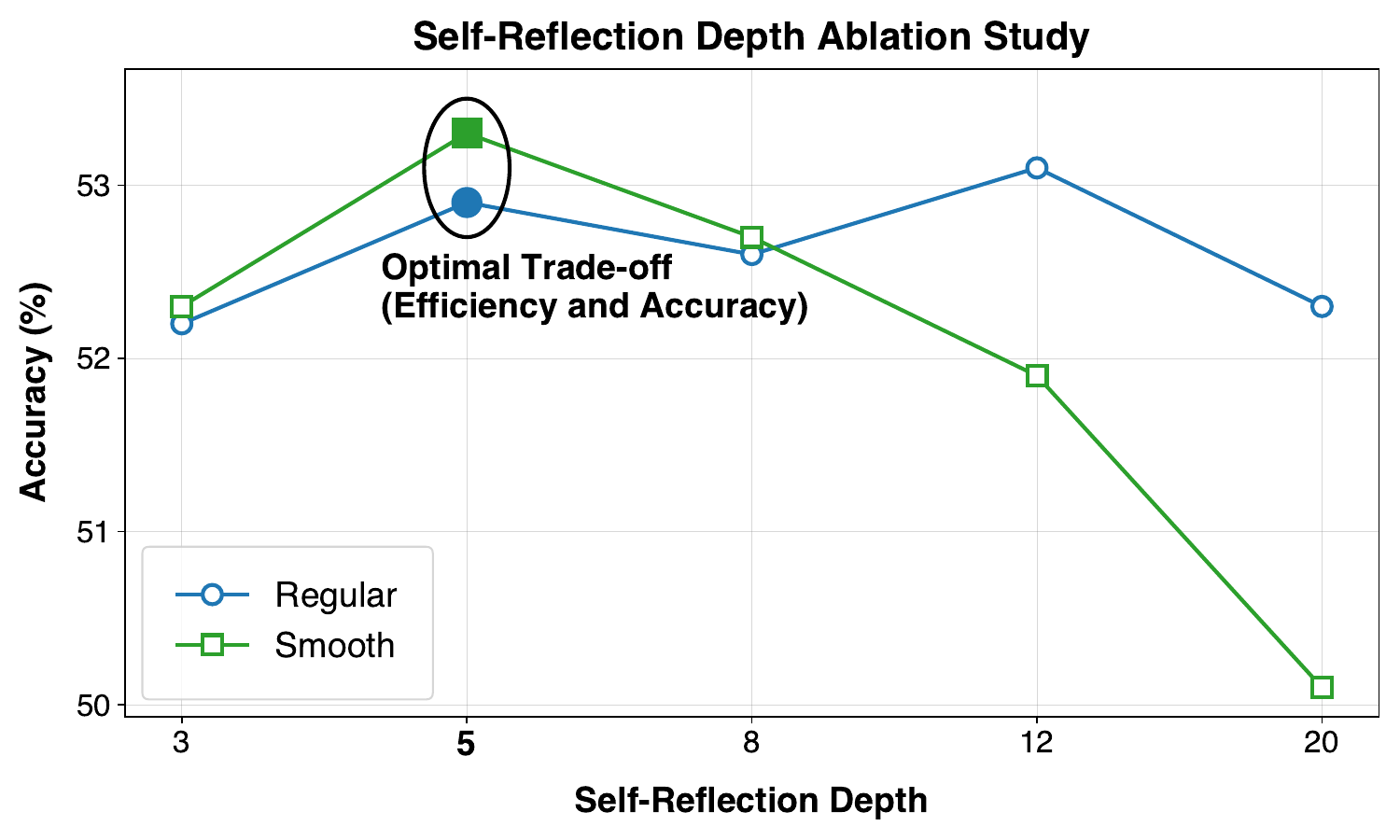}};
    \hspace{-5pt}
    % Place the table to the right
    \node[anchor=west] at (3.15, 0.15) {%
        \resizebox{!}{0.055\textwidth}{%
        \begin{tabular}{ccccc}
        \toprule
        $S$ & $T$ & $R$ & \multicolumn{2}{c}{Video-MME} \\ 
            &   &     & Long & All \\ 
        \midrule
        \multirow{3}{*}{32} & 64 & 2 &  53.3 & 63.2 \\
        & 96 & 3 & 53.9 & 63.6 \\
        & 128 & 4 & 54.6 & 64.3 \\ 
%        & 160 & 5 & 52.4 & 63.4 \\
        \midrule
        \multirow{1}{*}{64} &  128 & 2 & 56.3 & 65.5 \\
%        & 192 &  3 & 52.3 & 63.9 \\
        \bottomrule
        \end{tabular}
        }
    };
\end{tikzpicture}
\caption{\textbf{Ablation study on self-reflection layer depth (left) and context size width (right).} Experiments reveal that optimal self-reflection performance is achieved in earlier layers. Emulating larger context sizes with increased $T$ values also benefits accuracy.}
\label{fig:image_table}
\vspace{-10pt}
\end{figure}

This section analyzes the impact of \modelname's parameters on accuracy. Figure \ref{fig:image_table} reveals the effect of the depth of the self-reflection layer in the performance of \modelname~over long videos, using the Video-MME (Long)~\cite{VideoMME} benchmark. For consistency, we fix the segment size $S$ to $32$ and the total number of sampled frames $T$ to $64$.

We focus our ablation study in the shallow layers of the LLM. Considering the role of different layers in the forward pass of an LLM, early layers typically capture syntactic or input-level features, which are better for predictive tasks~\cite{fantastic} such as our designated spatiotemporal self-reflection, aiming to predict scores for visual tokens. We leave most deeper layers of the LLM out of our study, as they provide semantic and abstract representations geared towards text generation. 

%Within the ablated shallow layers, we pose that selecting the self-reflection layer depth requires analyzing the optimal trade-off between sufficient understanding of the input video and efficient inference.

Our results demonstrate that the \textit{Regular Inference} strategy, where the model resets the forward pass after self-reflection, observes a steady behavior throughout different depths. However, our analysis reports peak accuracy for \textit{Smooth Inference} at depth 5, and a rapid decline in deeper layers, corroborating the claims in \cite{fantastic}. Since the forward pass continues after self-reflection, and the visual hidden embeddings from different reflection paths are concatenated, the hidden embeddings of earlier layers represent more input-specific representations, while later layers represent abstract features that are relative for the answer being generated for each reflection path, and therefore are both worse for predicting visual token scores and for path convergence. \textit{Regular Inference} is more robust to this factor, since the forward pass restarts from the first layer using input tokens.

We also conduct additional experiments in Video-MME (Long) using varied configurations of $S$ and $T$, with a fixed self-reflection layer depth of $5$ for optimal performance-efficiency trade-off. The results are provided in Table~\ref{fig:image_table} and indicate a steady gain in accuracy by emulating a larger context size of $T$ frames and consequently more reflection paths $R$. This experiment confirms that we can effectively emulate larger context sizes within the base model's default context window by performing non-linear sampling. 

%
%NEED FURTHER JUSTIFICATION

\subsection{Computational Efficiency Analysis}

This section evaluates the impact of \modelname~on the inference speed of the baseline LLaVA-Video~\cite{LLaVA-Video} model. We measure these metrics on 70 videos from the Video-MME~\cite{VideoMME} benchmark, after a warm-up round of 30 videos, using a setup of 16GB of RAM and one NVIDIA A100 GPU with 40GB of VRAM.

\begin{figure}
    \centering
    \includegraphics[width=\linewidth, trim={0 12pt 0 0},clip]{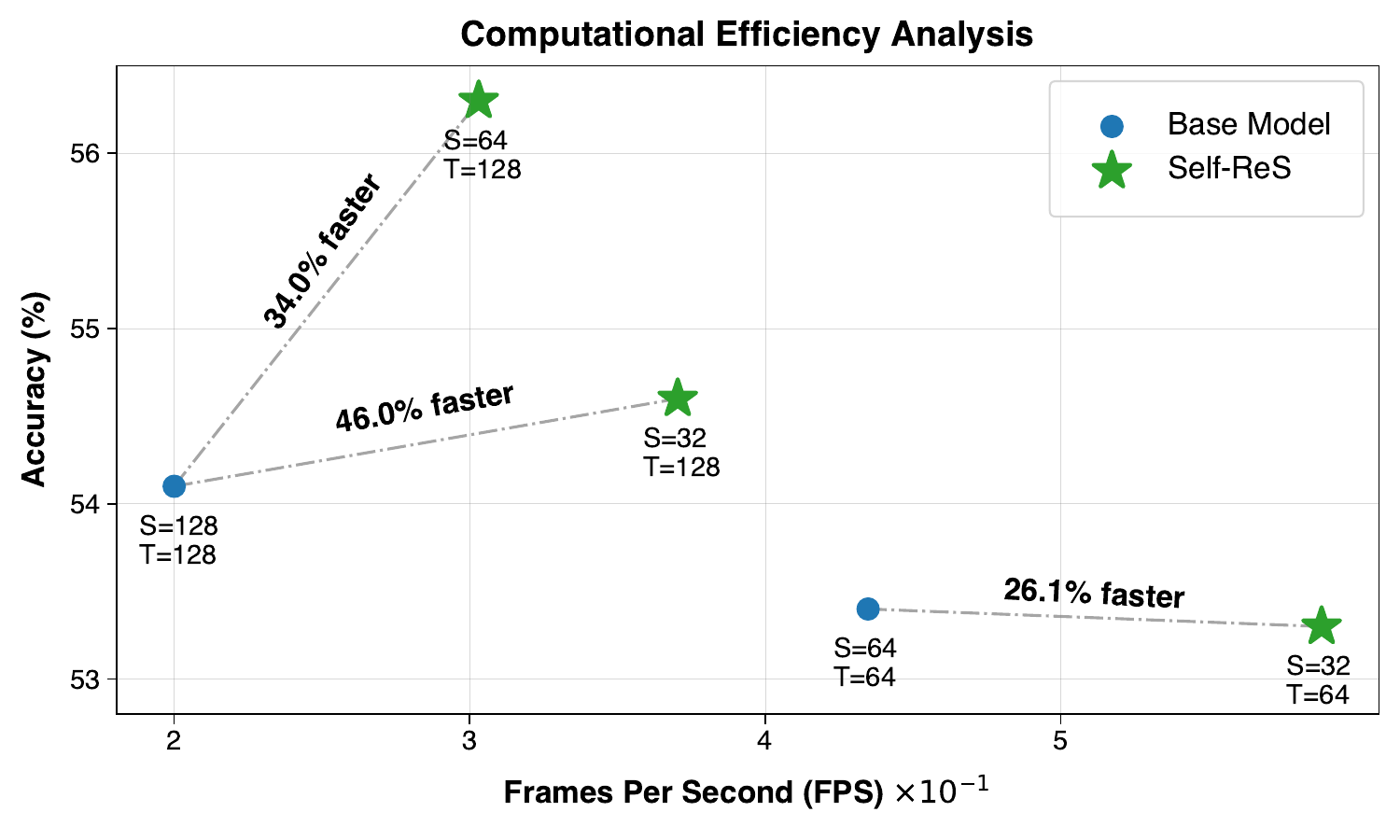}
    \caption{\textbf{Comparison of accuracy and inference speed on Video-MME for base model with and without \modelname, across varying context sizes.} Results show evidence that \modelname~has a positive effect on both the efficiency and accuracy of the base model.}
    \label{fig:inf_mem}
    \vspace{-1em}
\end{figure}

Figure \ref{fig:inf_mem} plots the accuracy in Video-MME (Long) and frames per second (fps) for \modelname~and the base model across different context sizes. For \modelname, we fix the segment size $S$ to 32 and 64, and vary the number of reflection paths $R$, emulating a larger context size of $T$ frames. We use \modelname~in batch mode, ensuring a fair comparison to the baseline. We purposely omit GPU memory consumption since we observed similar values when using batch mode. For information on inference speed and GPU memory consumption when using the memory efficient sequential mode, we refer the reader to the supplementary material.

We observe up to $\mathbf{46}$\textbf{\%} faster inference speed and  $\mathbf{2.2}$\textbf{\%} higher accuracy compared to the base model when using an emulated context size of $T=128$. As a result of our non-linear spatiotemporal sampling method, our solution can compress task-relevant information into a much smaller context size. Across all setups, \modelname~achieves similar or superior accuracy in faster inference speed.

\subsection{Qualitative Evaluation}
Our proposed spatiotemporal self-reflection process, guided by the user prompt, can effectively focus on the small spatiotemporal window where the main event occurs. In Figure~\ref{fig:teaser}, we see that \modelname~allows the model to correctly infer that the man blowtorches the door handles, despite this event happening in a small spatial and temporal window of the video. Other sampling strategies that perform linear sampling can introduce noise - e.g., frames that are not related to the man or the door handles - or miss key details - e.g., frames where the man interacts with the door lock and not the handles.
%Therefore, \modelname~captures relevant information for answering the user query.

%no texto sao equivalentes 

\section{Conclusion}
In this paper, we introduce \textbf{\modelname}, a novel self-reflective sampling method tailored to address the challenges of long-video understanding in LVLMs. By leveraging the inherent self-reflective abilities of LLMs, \modelname~selects key spatiotemporal fragments guided by an user query, effectively addressing video non-linearity without requiring additional training or external modules. Experiments on Video-MME and MLVU benchmarks demonstrate that \modelname~achieves state-of-the-art results among mid-sized LVLMs, surpassing models with significantly larger context sizes while reducing base model inference speed by up to \textbf{46\%}. \modelname~can be seamlessly integrated into existing LVLM architectures, enabling more efficient and accurate video understanding.
In future work, we plan to assess whether \modelname~can cope with augmented textual contexts and elicit chain-of-thought reasoning.

%\columnbreak

\vspace{-1em}
\bibliographystyle{IEEEbib}
\bibliography{icme2025references}

\begin{thebibliography}{10}

\bibitem{siglip}
Xiaohua Zhai, Basil Mustafa, and Alexander et~al. Kolesnikov,
\newblock ``Sigmoid loss for language image pre-training,''
\newblock in {\em ICCV}, 2023.

\bibitem{qwen2}
An~Yang, Baosong Yang, Binyuan Hui, Bo~Zheng, and et.al,
\newblock ``Qwen2 technical report,''
\newblock {\em arXiv:2407.10671}, 2024.

\bibitem{LLaVA-Video}
Yuanhan Zhang, Jinming Wu, Wei Li, Bo~Li, Zejun Ma, Ziwei Liu, and Chunyuan Li,
\newblock ``Video instruction tuning with synthetic data,''
\newblock {\em arXiv:2410.02713}, 2024.

\bibitem{VideoMME}
Chaoyou Fu, Yuhan Dai, and Yondong et~al. Luo,
\newblock ``Video-mme: The first-ever comprehensive evaluation benchmark of multi-modal llms in video analysis,''
\newblock {\em arXiv:2405.21075}, 2024.

\bibitem{Qwen2-VL}
Peng Wang, Shuai Bai, Sinan Tan, Shijie Wang, Zhihao Fan, and Jinze~Bai et~al.,
\newblock ``Qwen2-vl: Enhancing vision-language model's perception of the world at any resolution,''
\newblock {\em arXiv:2409.12191}, 2024.

\bibitem{NVILA}
Zhijian Liu, Ligeng Zhu, Baifeng Shi, Zhuoyang Zhang, Yuming Lou, and Shang Yang,
\newblock ``Nvila: Efficient frontier visual language models,''
\newblock {\em arXiv:2412.04468}, 2024.

\bibitem{VideoChatGPT}
Muhammad Maaz, Hanoona Rasheed, Salman Khan, and Khan et~al.,
\newblock ``Video-chatgpt: Towards detailed video understanding via large vision and language models,''
\newblock in {\em ACL}, 2024.

\bibitem{ChatUniVi}
Peng Jin, Ryuichi Takanobu, Wancai Zhang, and et~al. Cao,
\newblock ``Chat-univi: Unified visual representation empowers large language models with image and video understanding,''
\newblock in {\em CVPR}, 2024.

\bibitem{MovieChat}
Enxin Song, Wenhao Chai, Guanhong Wang, and et~al. Zhang,
\newblock ``Moviechat: From dense token to sparse memory for long video understanding,''
\newblock in {\em CVPR}, 2024.

\bibitem{LongVU}
Xiaoqian Shen, Yunyang Xiong, Changsheng Zhao, Lemeng Wu, Jun Chen, and Chenchen~Zhu et~al.,
\newblock ``Longvu: Spatiotemporal adaptive compression for long video-language understanding,''
\newblock {\em arXiv:2410.17434}, 2024.

\bibitem{oryx}
Zuyan Liu, Yuhao Dong, Ziwei Liu, and Hu,
\newblock ``Oryx mllm: On-demand spatial-temporal understanding at arbitrary resolution,''
\newblock {\em arXiv:2409.12961}, 2024.

\bibitem{VideoXL}
Yan Shu, Peitian Zhang, and Zheng Liu,
\newblock ``Video-xl: Extra-long vision language model for hour-scale video understanding,''
\newblock {\em arXiv:2409.14485}, 2024.

\bibitem{selfreflectionrag}
Akari Asai, Zeqiu Wu, Yizhong Wang, Avirup Sil, and Hannaneh Hajishirzi,
\newblock ``Self-{RAG}: Learning to retrieve, generate, and critique through self-reflection,''
\newblock in {\em ICLR}, 2024.

\bibitem{FastV}
Liang Chen, Haozhe Zhao, Tianyu Liu, Shuai Bai, Junyang Lin, Chang Zhou, and Baobao Chang,
\newblock ``An image is worth 1/2 tokens after layer 2: Plug-and-play inference acceleration for large vision-language models,''
\newblock in {\em ECCV}, 2024.

\bibitem{LLaVA}
Haotian Liu, Chunyuan Li, Qingyang Wu, and Yong~Jae Lee,
\newblock ``Visual instruction tuning,''
\newblock in {\em NeurIPS}, 2023.

\bibitem{BLIP2}
Junnan Li, Dongxu Li, Silvio Savarese, and Hoi et~al.,
\newblock ``{BLIP}-2: Bootstrapping language-image pre-training with frozen image encoders and large language models,''
\newblock in {\em ICML}, 2023.

\bibitem{li2024llamavid}
Yanwei Li, Chengyao Wang, and Jiaya Jia,
\newblock ``Llama-vid: An image is worth 2 tokens in large language models,''
\newblock 2024.

\bibitem{flash-attn}
Tri Dao, Daniel~Y. Fu, Stefano Ermon, Atri Rudra, and Christopher Ré,
\newblock ``Flashattention: Fast and memory-efficient exact attention with io-awareness,''
\newblock {\em arXiv:2205.14135}, 2022.

\bibitem{mlvu}
Junjie Zhou, Yan Shu, Bo~Zhao, Boya Wu, Shitao Xiao, and Xi~Yang et~al.,
\newblock ``Mlvu: A comprehensive benchmark for multi-task long video understanding,''
\newblock {\em arXiv:2406.04264}, 2024.

\bibitem{transformers}
Thomas Wolf, Lysandre Debut, Victor Sanh, Julien Chaumond, Clement Delangue, and Anthony~Moi et. al.,
\newblock ``Huggingface's transformers: State-of-the-art natural language processing,''
\newblock {\em arXiv:1910.03771}, 2020.

\bibitem{GPT4o}
OpenAI,
\newblock ``Gpt-4o,'' 2024,
\newblock Accessed: 2024-09-07.

\bibitem{GPT4o-mini}
OpenAI,
\newblock ``Gpt-4o mini,'' 2024,
\newblock Accessed: 2024-12-16.

\bibitem{VideoLLaMA2}
Zesen Cheng, Sicong Leng, Hang Zhang, Yifei Xin, Xin Li, Guanzheng Chen, Yongxin Zhu, Wenqi Zhang, Ziyang Luo, Deli Zhao, and Lidong Bing,
\newblock ``Videollama 2: Advancing spatial-temporal modeling and audio understanding in video-llms,''
\newblock {\em arXiv:2406.07476}, 2024.

\bibitem{LLaVA-OneVision}
Bo~Li, Yuanhan Zhang, Dong Guo, Renrui Zhang, Feng Li, Hao Zhang, Kaichen Zhang, Yanwei Li, Ziwei Liu, and Chunyuan Li,
\newblock ``Llava-onevision: Easy visual task transfer,''
\newblock {\em arXiv:2408.03326}, 2024.

\bibitem{fantastic}
Zhu Liu, Cunliang Kong, and Ying et~al. Liu,
\newblock ``Fantastic semantics and where to find them: Investigating which layers of generative llms reflect lexical semantics,''
\newblock {\em arXiv:2403.01509}, 2024.

\end{thebibliography}

% \vspace{12pt}
% \color{red}
% IEEE conference templates contain guidance text for composing and formatting conference papers. Please ensure that all template text is removed from your conference paper prior to submission to the conference. Failure to remove the template text from your paper may result in your paper not being published.

\appendix

\subsection{Reconstructing Context After Convergence}
\label{app:pos_embbs}

We reconstruct the context using only the most relevant visual tokens, reassigning position embeddings as a sequential 1D index. Our strategy shifts the positions of tokens beyond the first reflection path without disrupting their spatial-temporal structure. LVLMs process inputs as 1D token sequences, and spatial structure is only enforced by the vision encoder, which remains unchanged. Although token distances may be altered, their relative order is maintained, ensuring coherence in the final representation.
We validate our approach using our best setup against three alternatives:
1) \textbf{Retaining original positional embeddings} – Results in a non-incremental sequence, duplicating positions for tokens from different reflection paths, and introduces varying distances between tokens; 2) \textbf{Retaining original positional embeddings incremented according to their reflection path} – Avoids duplicate tokens but introduces varying distances between tokens; 3) \textbf{Using a single position embedding} – Removes temporal position information from vision tokens. As shown in Table~\ref{tab:pos_embbs}, our original strategy of reassigning positional embeddings yields the best performance.
\begin{table}[!h]
\centering
\caption{Position Embedding Assignment Strategy (Performance on Video-MME).}
\label{tab:pos_embbs}
%\setlength{\tabcolsep}{4pt} % Reduces column padding
%\renewcommand{\arraystretch}{1.1} % Adjusts row height
%\begin{adjustbox}{max width=0.45\columnwidth}
\begin{tabular}{lcc}
    \toprule
    \textbf{Strategy} & \textbf{Long} & \textbf{Overall} \\
    \midrule
    1) & 51.8 & 62.1 \\
    2) & 52.3 & 62.3 \\
    3) & 41.7 & 50.9 \\
    \midrule
    Ours & \textbf{54.6} & \textbf{64.3} \\
    \bottomrule
\end{tabular}
%\end{adjustbox}
\end{table}

\subsection{Sequential Mode}

This section evaluates the impact of \modelname~on the GPU memory budget and inference speed of the base LLaVA-Video~\cite{LLaVA-Video} model. We measure these metrics on 70 videos from the Video-MME~\cite{VideoMME} benchmark, after a warm-up round of 30 videos, using the same setup used in batch mode, with 16GB of RAM and one NVIDIA A100 GPU with 40GB of VRAM.

\begin{figure}[t]
    \centering
    \includegraphics[width=\linewidth]{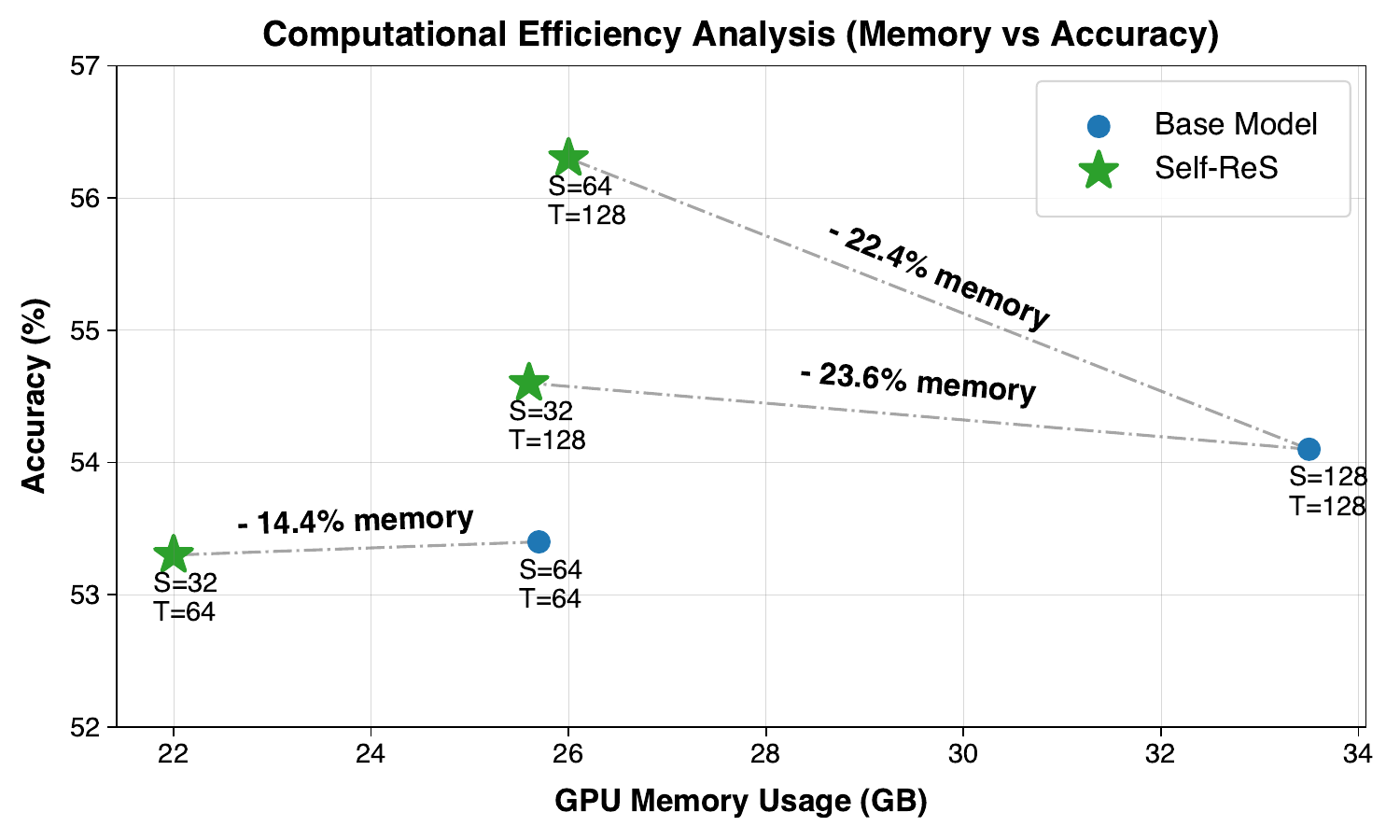}
    \caption{\textbf{GPU Memory vs Accuracy.}}
    \label{fig:gpu}
\end{figure}

Figure \ref{fig:gpu} plots the accuracy in Video-MME (Long) and maximum GPU memory consumption during inference for \modelname~and the base model across different context sizes. For \modelname, we fix the segment size $S$ to 32 and 64, and vary the number of reflection paths $R$, emulating a larger context size of $T$ frames. We use \modelname~in sequential mode, and split the reflection paths over two mini-batches. During spatiotemporal self-reflection, \modelname~processes each mini batch sequentially in the GPU, storing the remaining mini-batch in RAM. After convergence into a single path, \modelname~conducts inference in GPU only.  

\begin{figure}[t]
    \centering
    \includegraphics[width=\linewidth]{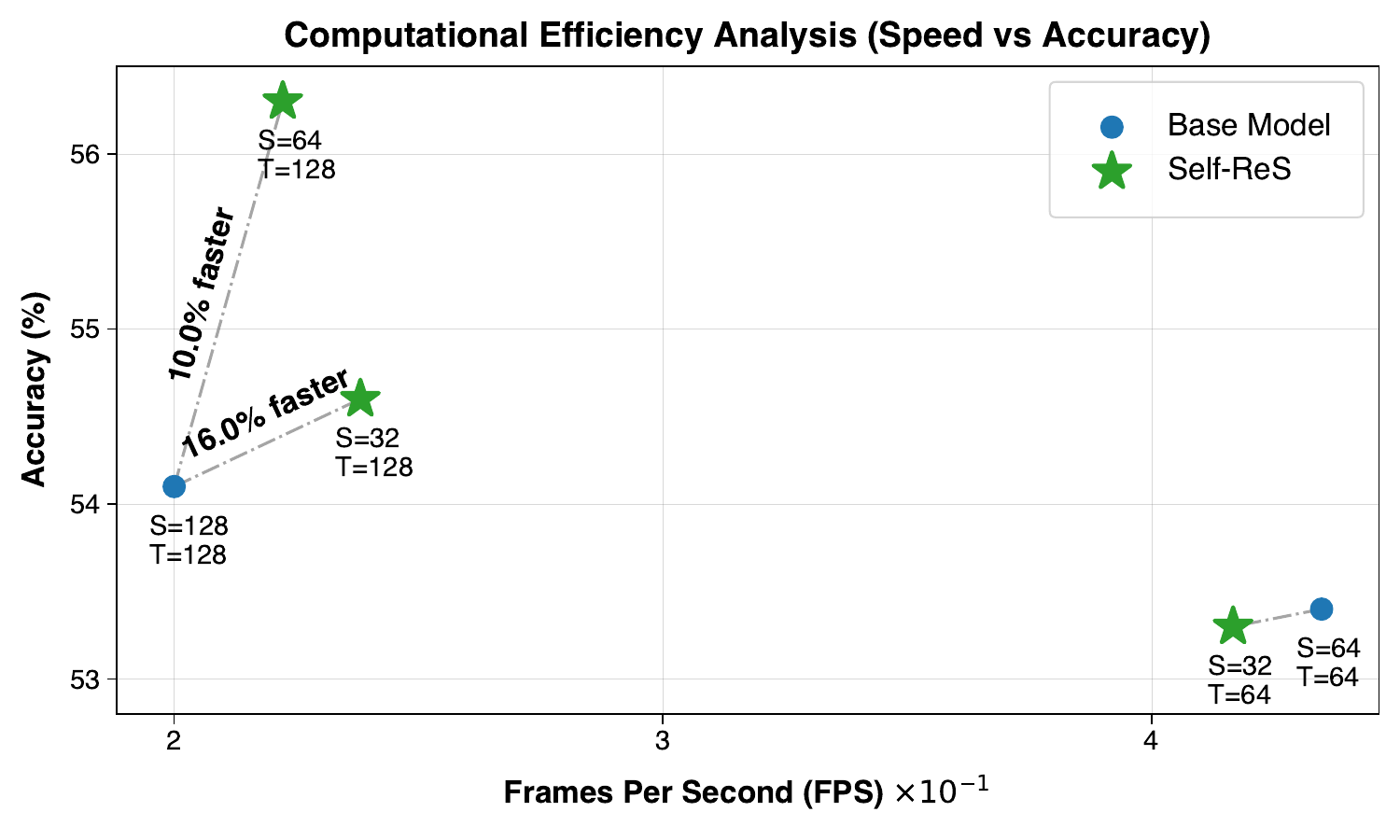}
    \caption{\textbf{Inference Speed vs Accuracy.}}
    \label{fig:inf}
\end{figure}

Results show \modelname~reduces the maximum GPU memory consumption by up to $\mathbf{23.6}$\textbf{\%}. Notably, despite sequential processing, our approach is still $\mathbf{16}$\textbf{\%} faster than the base model for a context of $T=128$ - Figure \ref{fig:inf}. Therefore, \modelname~allows faster inference over emulated larger context sizes in restrained GPU memory environments.


\subsection{Expanding Baselines}
We extend \modelname~to LLaVA-OneVision~\cite{LLaVA-OneVision} and show that our approach has a positive impact on performance (Table ~\ref{tab:table2}).

\begin{table}[h]
\centering
\caption{Performance of \modelname~extended LLaVA-OneVision (VideoMME).}
\label{tab:table2}
%\setlength{\tabcolsep}{4pt} % Reduces column padding
%\renewcommand{\arraystretch}{1.1} % Adjusts row height
%\begin{adjustbox}{max width=0.6\columnwidth}
\begin{tabular}{lcccc}
    \toprule
    \textbf{Model} & \textbf{Size} & \textbf{Long} & \textbf{Overall} \\
    \midrule
    LLaVA-OV~\cite{LLaVA-OneVision} & \multirow{2}{*}{7B} & 48.6 & 58.2\\
    w/ \modelname & & \textbf{50.7} &\textbf{59.1} \\
    %\midrule
    %VideoLLaMA3 & \multirow{2}{*}{7B} & 66.2  \\
    %w/ Self-ReS & & \\
    \bottomrule
\end{tabular}
%\end{adjustbox}
\end{table}
\subsection{Additional Dataset Details}
\subsubsection{Video-MME}
The Video-MME~\cite{VideoMME} dataset contains 900 videos with a total of 254 hours. The videos, ranging from 11 seconds to 1 hour, were manually selected and annotated by human experts, generating a total of 2,700 question-answer pairs. Questions are split across the following categories: Object Recognition, Action Recognition, OCR Problems, Counting Problems, Temporal Reasoning, Spatial Reasoning, Action Reasoning, Object Reasoning, Information Synopsis, Temporal Perception, Spatial Perception and Attribute Perception.
\subsubsection{MLVU}
The MLVU~\cite{mlvu} dataset contains a total of 1334 instances of long videos ranging from 3 minutes to 2 hours. The dataset contains 2593 questions, split into 2175 multiple-choice and 418
free-form generation questions. These question were either automatically generated based on previously existing annotations --e.g., Anomaly Recognition category- or manually annotated by human experts.

\subsection{Extended Performance on all Video Lengths}
\modelname~ delivers superior performance across different video lengths. Table \ref{tab:video_lengths} shows that \modelname~improves performance for medium and long videos. In short videos, performance is maintained since most short videos require a single reflection path, resulting in the same setup as the base model.

\begin{table}[h]
\centering
\caption{Performance across video lengths (VideoMME).}
%\begin{adjustbox}{max width=\columnwidth}
\begin{tabular}{lccccc}
\toprule
\multirow{2}{*}{\textbf{Model}} & \multirow{2}{*}{\textbf{T}} & \textbf{Short} & \textbf{Medium} & \textbf{Long} & \multirow{2}{*}{\textbf{Overall}} \\ 
 & &  $<2$min &  4-15 min &  30-60 min & \\ \midrule

\multirow{2}{*}{Baseline} &32 & 76.3  & 59.0   & 52.2 & 62.5    \\
& 64   & 77.2  & 62.1   & 53.4 & 64.3    \\\midrule
%& 128 & 76.3  & 63.4   & 54.1 & 64.6    \\ \midrule

\multirow{2}{*}{Self-ReS}& 128 (32) & 76.2  & 62.0   & 54.6 & 64.3    \\
& 128 (64) & \textbf{77.2}  & \textbf{63.7}   & \textbf{56.3} & \textbf{65.5}    \\ \bottomrule
\end{tabular}
%\end{adjustbox}
\label{tab:video_lengths}
\end{table}

\subsection{Extra Qualitative Examples}

\begin{figure*}[b]
    \centering
    \includegraphics[width=\textwidth]{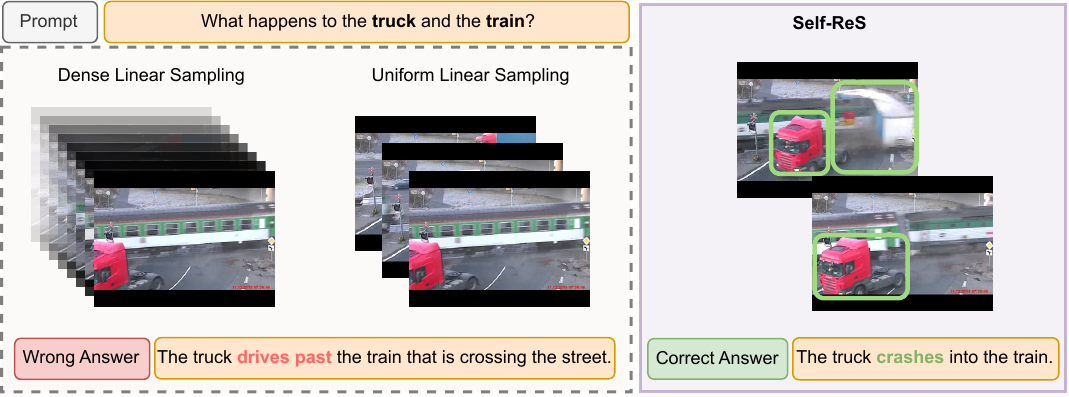}
    \caption{\textbf{Linear Sampling vs \modelname.} With \modelname, the model can detect the crash.}
    \label{fig:qualitative}
\end{figure*}

\begin{figure*}[b]
    \centering
    \includegraphics[width=\textwidth]{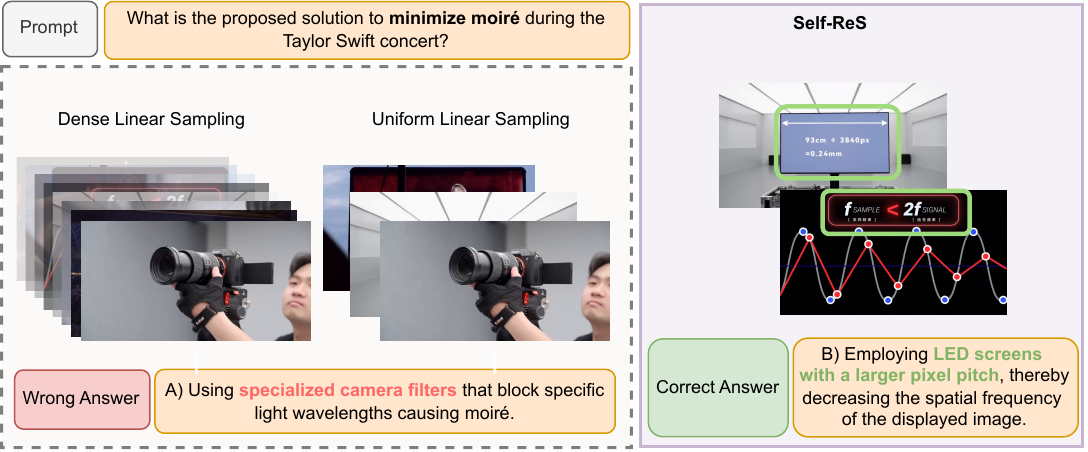}
    \caption{\textbf{Linear Sampling vs \modelname.} With \modelname, the model can focus on the visual cue indicating led screens with larger pixel pitch.}
    %Linear sampling introduces redundant and background information, missing the small spatiotemporal windows essential for answering the question.}
    \label{fig:qualitative2}
\end{figure*}

Figures \ref{fig:qualitative} and \ref{fig:qualitative2} provide additional visual examples of the effectiveness of the proposed approach. In Figure \ref{fig:qualitative}, the base sampling strategy misses the small spatial and temporal window where the road accident occurs. In Figure \ref{fig:qualitative2}, the base sampling strategy samples a background picture of a camera, leading the model towards an incorrect answer. With \modelname~integration, the model can leverage textual cues present in the user prompt and reason over key spatiotemporal fragments of the video that will guide the model towards the correct answer.

%With \modelname, the model is able to self-reflect on the prompt and focus on a small spatiotemporal window of the video with important visual fragments that better align with the question, generating the correct answer. 


\end{document}